\documentclass{article}

\usepackage[final]{neurips_2019}

\usepackage[utf8]{inputenc} 
\usepackage[T1]{fontenc}    
\usepackage{hyperref}       
\usepackage{url}            
\usepackage{booktabs}       
\usepackage{amsfonts}       
\usepackage{nicefrac}       
\usepackage{microtype}      
\usepackage{amsmath}
\usepackage{graphicx}
\usepackage{subfigure}
\usepackage{verbatim}
\usepackage{bm}
\usepackage[dvipsnames]{xcolor}
\usepackage{wrapfig}
\usepackage{adjustbox}
\usepackage{footnote}

\newcommand{\method}{GRU-ODE-Bayes}

\newcounter{daggerfootnote}
\newcommand*{\daggerfootnote}[1]{%
    \setcounter{daggerfootnote}{\value{footnote}}%
    \renewcommand*{\thefootnote}{\fnsymbol{footnote}}%
    \footnote[2]{#1}%
    \setcounter{footnote}{\value{daggerfootnote}}%
    \renewcommand*{\thefootnote}{\arabic{footnote}}%
    }

\DeclareMathOperator{\gruode}{\mathbf{GRU-ODE}}
\DeclareMathOperator{\grubayes}{\mathbf{GRU-Bayes}}
\DeclareMathOperator{\gru}{\mathbf{GRU}}
\DeclareMathOperator{\loss}{Loss}

\title{\method: Continuous modeling of sporadically-observed time series}

\author{
  Edward~De Brouwer\thanks{Both authors contributed equally}  \thanks{Corresponding author} \\
  ESAT-STADIUS\\
  KU LEUVEN\\
  Leuven, 3001, Belgium \\
  \texttt{edward.debrouwer@esat.kuleuven.be} \\
  \And
  Jaak Simm\footnotemark[1] \\
  ESAT-STADIUS \\
  KU LEUVEN \\
  Leuven, 3001, Belgium  \\
  \texttt{jaak.simm@esat.kuleuven.be}\\
  \AND
  Adam Arany \\
  ESAT-STADIUS \\
  KU LEUVEN \\
  Leuven, 3001, Belgium\\
   \texttt{adam.arany@esat.kuleuven.be}\\
  \And
  Yves Moreau \\
   ESAT-STADIUS \\
   KU LEUVEN \\
  Leuven, 3001, Belgium\\
  \texttt{moreau@esat.kuleuven.be}\\
}

\begin{document}

\maketitle

\begin{abstract}
Modeling real-world multidimensional time series can be particularly challenging when these are \emph{sporadically} observed (\emph{i.e.}, sampling is irregular both in time and across dimensions)---such as in the case of clinical patient data. To address these challenges, we propose (1) a continuous-time version of the Gated Recurrent Unit, building upon the recent Neural Ordinary Differential Equations \citep{neural_ode}, and (2) a Bayesian update network that processes the sporadic observations. We bring these two ideas together in our \method{} method. We then demonstrate that the proposed method encodes a continuity prior for the latent process and that it can exactly represent the Fokker-Planck dynamics of complex processes driven by a multidimensional stochastic differential equation. 
Additionally, empirical evaluation shows that our method outperforms the state of the art on both synthetic data and real-world data with applications in healthcare and climate forecast. What is more, the continuity prior is shown to be well suited for low number of samples settings.

\end{abstract}

\section{Introduction}

Multivariate time series are ubiquitous in various domains of science, such as healthcare \citep{jensen2014temporal}, astronomy \citep{scargle1982studies}, 
or climate science \citep{schneider2001analysis}. Much of the methodology for time-series analysis assumes that signals are measured systematically at fixed time intervals. 
However, much real-world data can be \emph{sporadic} (\emph{i.e.}, the signals are sampled irregularly and not all signals are measured each time). A typical example is patient measurements, which are taken when the patient comes for a visit (\emph{e.g.,} sometimes skipping an appointment) and where not every measurement is taken at every visit. Modeling then becomes challenging as such data violates the main assumptions underlying traditional machine learning methods (such as recurrent neural networks). 

Recently, the Neural Ordinary Differential Equation (ODE) model \citep{neural_ode} opened the way for a novel, continuous representation of neural networks. As time is intrinsically continuous, this framework is particularly attractive for time-series analysis. It opens the perspective of tackling the issue of irregular sampling in a natural fashion, by integrating the dynamics over whatever time interval needed. Up to now however, such ODE dynamics have been limited to the continuous \emph{generation} of observations (\emph{e.g.}, decoders in variational auto-encoders (VAEs) \citep{kingma2013auto} or normalizing flows~\citep{rezende2014stochastic}). 

Instead of the encoder-decoder architecture where the ODE part is decoupled from the input processing, we introduce a tight integration by \emph{interleaving} the ODE and the input processing steps. Conceptually, this allows us to drive the dynamics of the ODE directly by the incoming sporadic inputs. To this end, we propose (1) a continuous time version of the Gated Recurrent Unit and (2) a Bayesian update network that processes the sporadic observations. We combine these two ideas to form the \method{} method.

\begin{figure}[tbh]
\vskip 0.0in
\begin{center}
\centerline{\includegraphics[width=10cm]{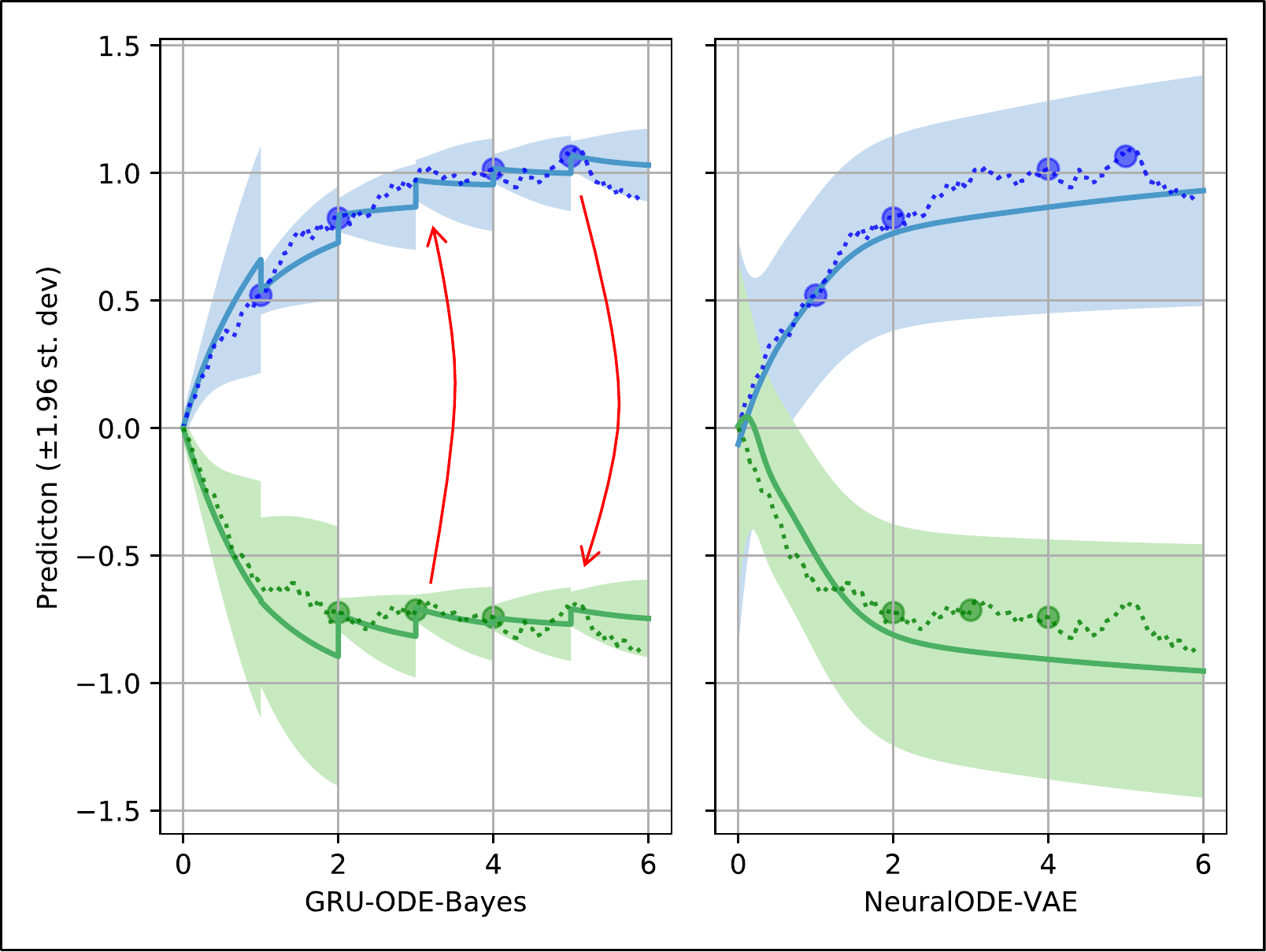}}
\caption{Comparison of GRU-ODE-Bayes and NeuralODE-VAE on a 2D Ornstein-Uhlenbeck process with highly correlated Wiener processes ($\rho=0.99)$. Dots are the values of the actual underlying process (dotted lines) from which the sporadic observations are obtained. Solid lines and shaded areas are the inferred means and 95\% confidence intervals. Note the smaller errors and smaller variance of GRU-ODE-Bayes vs. NeuralODE-VAE. Note also that GRU-ODE-Bayes can infer that a jump in one variable also implies a jump in the other unobserved one (red arrows). Similarly, it also learns the reduction of variance resulting from a new incoming observation.}
\label{fig:comparison}
\end{center}
\vskip -0.2in
\end{figure}

The tight coupling between observation processing and ODE dynamics allows the proposed method to model fine-grained nonlinear dynamical interactions between the variables. As illustrated in Figure~\ref{fig:comparison}, \method{} can (1) quickly infer the unknown parameters of the underlying stochastic process and (2) learn the correlation between its variables (red arrows in Figure~\ref{fig:comparison}). 
In contrast, the encoder-decoder based method NeuralODE-VAE proposed by \citet{neural_ode} captures the general structure of the process without being able to recover detailed interactions between the variables (see Section~\ref{sec:SyntheticSDEs} for detailed comparison).

Our model enjoys important theoretical properties. We frame our analysis in a general way by considering that observations follow the dynamics driven by a stochastic differential equation (SDE). 
In Section~\ref{sec:SyntheticSDEs} and Appendix~\ref{app:OU_SDE}, we show that \method{} can exactly represent the corresponding Fokker-Planck dynamics in the special case of the Ornstein-Uhlenbeck process, as well as in generalized versions of it. 
We further perform an empirical evaluation and show that our method outperforms the state of the art on healthcare and climate data (Section~\ref{sec:real_world}).

\subsection{Problem statement}
We consider the general problem of forecasting on $N$ sporadically observed \(D\)-dimensional time series. For example, data from \(N\) patients where \(D\) clinical longitudinal variables can potentially be measured. Each time series $i \in \{1,\ldots,N\}$ is measured at \(K_i\) time points specified by a vector of observation times $\mathbf{t}_i \in \mathbb{R}^{K_i}$. 
The values of these observations are specified by a matrix of observations $\mathbf{y}_i \in \mathbb{R}^{K_i \times D}$ and an observation mask $\mathbf{m}_i \in \{0,1\}^{K_i\times D}$ (to indicate which of the variables are measured at each time point). 
 
We assume that observations $\mathbf{y}_i$ are sampled from the realizations of a $D$-dimensional stochastic process $\mathbf{Y}(t)$ whose dynamics is driven by an unknown SDE: 
\begin{align}
d\mathbf{Y}(t) = \mu(\mathbf{Y}(t))dt + \sigma(\mathbf{Y}(t))d\mathbf{W}(t),
\label{eq:SDE}
\end{align}
where $d\mathbf{W}(t)$ is a Wiener process. The distribution of $\mathbf{Y}(t)$ then evolves according to the celebrated Fokker-Planck equation \citep{risken1996fokker}. We refer to the mean and covariance parameters of its probability density function (PDF) as $\mu_{\mathbf{Y}}(t)$ and $\Sigma_{\mathbf{Y}}(t)$. 

Our goal will be to model the unknown temporal functions $\mu_{\mathbf{Y}}(t)$  and $\Sigma_{\mathbf{Y}}(t)$ from the sporadic measurements $\mathbf{y}_i$. These are obtained by sampling the random vectors $\mathbf{Y}(t)$ at times $\mathbf{t}_i$ with some observation noise $\bm{\epsilon}$. Not all dimensions are sampled each time, resulting in missing values in $\mathbf{y}_i$. 
In contrast to classical SDE inference \citep{sarkka2019applied}, we consider the functions $\mu_{\mathbf{Y}}(t)$ and $\Sigma_{\mathbf{Y}}(t)$ are parametrized by neural networks.

This SDE formulation is general. It embodies the natural assumption that seemingly identical processes can evolve differently because of unobserved information. In the case of intensive care, as developed in Section~\ref{sec:real_world}, it reflects the evolving uncertainty regarding the patient's future condition. 


\section{Proposed method}
At a high level, we propose a dual mode system consisting of (1) a GRU-inspired continuous-time state evolution (GRU-ODE) that \emph{propagates} in time the hidden state $\mathbf{h}$ of the system between observations and (2) a network that \emph{updates} the current hidden state to incorporate the incoming observations (GRU-Bayes). The system switches from propagation to update and back whenever a new observation becomes available.

We also introduce an observation model $f_{\mathrm{obs}}(\mathbf{h}(t))$ mapping $\mathbf{h}$ to the estimated parameters of the observations distribution $\mu_{\mathbf{Y}}(t)$  and $\Sigma_{\mathbf{Y}}(t)$ (details in Appendix~\ref{app:f_obs}). GRU-ODE then explicitly learns the Fokker-Planck dynamics of Eq.~\ref{eq:SDE}. This procedure allows end-to-end training of the system to minimize the loss with respect to the sporadically sampled observations $\mathbf{y}$.

\subsection{GRU-ODE derivation}
\label{sec:gru-ode}
To derive the GRU-based ODE, we first show that the GRU proposed by \citet{cho14:gru} can be written as a difference equation. First, let  $\mathbf{r}_t$, $\mathbf{z}_t$, and $\mathbf{g}_t$  be the reset gate, update gate, and update vector of the GRU: 
\begin{align}
  \mathbf{r}_t &= \sigma(W_r \mathbf{x}_t + U_r \mathbf{h}_{t-1}+\mathbf{b}_r) \nonumber \\
  \mathbf{z}_t &= \sigma(W_z \mathbf{x}_t + U_z \mathbf{h}_{t-1}+\mathbf{b}_z) 
  \label{eq:gru-original} \\
  \mathbf{g}_t &= \tanh(W_h  \mathbf{x}_t + U_h (\mathbf{r}_t \odot \mathbf{h}_{t-1}) +  \mathbf{b}_h), \nonumber
\end{align}
where \(\odot\) is the elementwise product. Then the standard update for the hidden state $\mathbf{h}$ of the GRU is
\begin{align*}
  \mathbf{h}_t &= \mathbf{z}_t \odot \mathbf{h}_{t-1} + (1 - \mathbf{z}_t) \odot \mathbf{g}_t.
\end{align*}
We can also write this as \(\mathbf{h}_t=\gru(\mathbf{h}_{t-1},\mathbf{x}_t)\). By subtracting $\mathbf{h}_{t-1}$ from this state update equation and factoring out $(1 - \mathbf{z}_t)$, we obtain a difference equation
\begin{align*}
	\Delta \mathbf{h}_t = \mathbf{h}_t - \mathbf{h}_{t-1}
	&= \mathbf{z}_t \odot \mathbf{h}_{t-1} + (1 - \mathbf{z}_t) \odot \mathbf{g}_t - \mathbf{h}_{t-1}
	\nonumber
	\\
    &=  (1 - \mathbf{z}_t) \odot (\mathbf{g}_t - \mathbf{h}_{t-1}).
\end{align*}
This difference equation naturally leads to the following ODE for $\mathbf{h}(t)$:
\begin{align}
  \label{eq:gru-ode-update}
  \dfrac{d \mathbf{h}(t)}{dt} =  (1 - \mathbf{z}(t)) \odot (\mathbf{g}(t) - \mathbf{h}(t)),
\end{align}

where $\mathbf{z}$, $\mathbf{g}$, $\mathbf{r}$ and $\mathbf{x}$ are the continuous counterpart of Eq.~\ref{eq:gru-original}. See Appendix~\ref{app:gru-ode-full} for the explicit form.

We name the resulting system \emph{GRU-ODE}. Similarly, we derive the \emph{minimal GRU-ODE}, a variant based on the minimal GRU~\citep{zhou2016minimal}, described in appendix \ref{app:gru-minimal}.

In case \emph{continuous observations} or \emph{control signals} are available, they can be naturally fed to the GRU-ODE input $\mathbf{x}(t)$. For example, in the case of clinical trials, the administered daily doses of the drug under study can be used to define a continuous input signal. If no continuous input is available, then nothing is fed as $\mathbf{x}(t)$ and the resulting ODE in Eq.~\ref{eq:gru-ode-update} is autonomous, with $\mathbf{g}(t)$ and $\mathbf{z}(t)$ only depending on $\mathbf{h}(t)$.

\subsection{General properties of GRU-ODE}
\label{subsec:gru_ode_properties}
GRU-ODE enjoys several useful properties:

\textbf{Boundedness.}~ First, the hidden state $\mathbf{h}(t)$ stays within the $[-1, 1]$ range\footnote{We use the notation $[-1, 1]$ to also mean multi-dimensional range (\emph{i.e.}, all elements are within $[-1, 1]$).}. This restriction is crucial for the compatibility with the GRU-Bayes model and comes from the negative feedback term in Eq.~\ref{eq:gru-ode-update}, which stabilizes the resulting system. 
In detail, if the $j$-th dimension of the starting state $\mathbf{h}(0)$ is within $[-1, 1]$, then $\mathbf{h}(t)_j$ will always stay within $[-1, 1]$ because
\begin{equation*}
    \left . \dfrac{d\mathbf{h}(t)_j}{dt} \right|_{t: \mathbf{h}(t)_j=1} \leq 0
    \quad\text{ and }\quad
    \left . \dfrac{d\mathbf{h}(t)_j}{dt} \right|_{t: \mathbf{h}(t)_j=-1} \geq 0.
\end{equation*}
This can be derived from the ranges of $\mathbf{z}$ and $\mathbf{g}$ in Eq.~\ref{eq:gru-original}. Moreover, would $\mathbf{h}(0)$ start outside of the $[-1, 1]$ region, the negative feedback would quickly push $\mathbf{h}(t)$ into this region, making the system also robust to numerical errors.

\textbf{Continuity.}~Second, GRU-ODE is Lipschitz continuous with constant $K=2$. Importantly, this means that GRU-ODE encodes a \emph{continuity prior} for the latent process $\mathbf{h}(t)$. This is in line with the assumption of a continuous hidden process generating observations (Eq.~\ref{eq:SDE}). In Section \ref{sec:continuity}, we demonstrate empirically the importance of this prior in the small-sample regime.

\textbf{General numerical integration.}~ As a parametrized ODE, GRU-ODE can be integrated with any numerical solver. In particular, adaptive step size solvers can be used. Our model can then afford large time steps when the internal dynamics is slow, taking advantage of the continuous time formulation of Eq.~\ref{eq:gru-ode-update}. It can also be made faster with sophisticated ODE integration methods. We implemented the following methods: Euler, explicit midpoint, and Dormand-Prince (an adaptive step size method). 
Appendix~\ref{app:dopri} illustrates that the Dormand-Prince method requires fewer time steps. 



\subsection{GRU-Bayes}
\label{subsec:gru_bayes}
GRU-Bayes is the module that processes the sporadically incoming observations to update the hidden vectors, and hence the estimated PDF of $\mathbf{Y}(t)$.
This module is based on a standard GRU and thus operates in the region $[-1, 1]$ that is required by GRU-ODE. In particular, GRU-Bayes is able to update $\mathbf{h}(t)$ to any point in this region. Any adaptation is then within reach with a \emph{single observation}.

To feed the GRU unit inside GRU-Bayes with a non-fully-observed vector, we first preprocess it with an observation mask using $f_{\mathrm{prep}}$, as described in Appendix~\ref{app:fprep}. For a given time series, the resulting update for its $k$-th observation $\mathbf{y}[k]$ at time $t=\mathbf{t}[k]$ with mask $\mathbf{m}[k]$ and hidden vector $\mathbf{h}(t_-)$ is
\begin{equation}
    \mathbf{h}(t_+) = \gru(\mathbf{h}(t_-), f_{\mathrm{prep}}(\mathbf{y}[k], \mathbf{m}[k], \mathbf{h}(t_-))),
    \label{eq:GRU-Bayes}
\end{equation}
where $\mathbf{h}(t_-)$ and $\mathbf{h}(t_+)$ denote the hidden representation before and after the jump from GRU-Bayes update. We also investigate an alternative option where the $\mathbf{h}(t)$ is updated by each observed dimension \emph{sequentially}. We call this variant \emph{GRU-ODE-Bayes-seq} (see Appendix~\ref{app:gru-seq} for more details). In Appendix~\ref{app:ablation_bayes}, we run an ablation study of the proposed GRU-Bayes architecture by replacing it with a MLP and show that the aforementioned properties are crucial for good performance.

\subsection{\method{}}
\begin{figure}[H]
\vskip 0.0in 
\begin{center}
\centerline{\includegraphics[width=7cm]{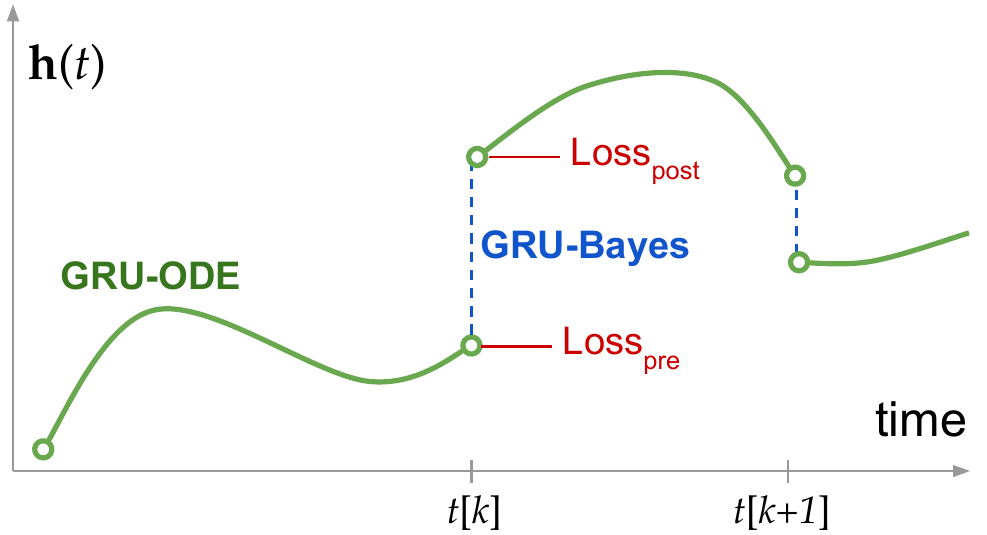}}
\caption{GRU-ODE-Bayes uses GRU-ODE to evolve the hidden state between two observation times $t[k]$ and $t[k+1]$. GRU-Bayes processes the observations and updates the hidden vector $\mathbf{h}$ in a discrete fashion, reflecting the additional information brought in by the observed data.}
\label{fig:gru-ode-bayes}
\end{center}
\vskip -0.2in
\end{figure}

The proposed \method{} combines GRU-ODE and GRU-Bayes. The GRU-ODE is used to evolve the hidden state $\mathbf{h}(t)$ in continuous time between the observations and GRU-Bayes transforms the hidden state, based on the observation $\mathbf{y}$, from $\mathbf{h}(t_-)$ to $\mathbf{h}(t_+)$. As best illustrated in Figure~\ref{fig:gru-ode-bayes}, the alternation between GRU-ODE and GRU-Bayes results in an ODE with \emph{jumps}, where the jumps are at the locations of the observations.

GRU-ODE-Bayes is best understood as a filtering approach. Based on previous observations (until time $t_k$), it can estimate the probability of future observations. Like the celebrated Kalman filter, it alternates between a \emph{prediction} (GRU-ODE) and a \emph{filtering} (GRU-Bayes) phase. Future values of the time series are predicted by integrating the hidden process $\mathbf{h}(t)$ in time, as shown on the green solid line in Figure \ref{fig:gru-ode-bayes}. The update step discretely updates the hidden state when a new measurement becomes available (dotted blue line). Let's note that unlike the Kalman filter, our approach is able to learn complex dynamics for the hidden process.

\subsubsection*{Objective function}
To train the model using sporadically-observed samples, we introduce two losses. The first loss, $\loss_{\mathrm{pre}}$, is computed before the observation update and is the negative log-likelihood (NegLL) of the observations. For the observation of a single sample, we have (for readability we drop the time indexing):
\begin{equation*}
    \loss_{\mathrm{pre}}
    = - \sum_{j=1}^D m_j \log p(y_j | \theta=f_{\mathrm{obs}}(\mathbf{h}_-)_j),
\end{equation*}
where $m_j$ is the observation mask and $f_{\mathrm{obs}}(\mathbf{h}_-)_j$ are the parameters of the distribution before the update, for dimension $j$. Thus, the error is only computed on the observed values of $\mathbf{y}$.

For the second loss, let $p_{\mathrm{pre}}$ denote the predicted distributions (from $\mathbf{h}_-$) before GRU-Bayes.

\begin{wrapfigure}[18]{r}{0.5\textwidth}
\begin{minipage}{0.5\textwidth}
\begin{algorithm}[H]
   \caption{GRU-ODE-Bayes}
   \label{alg:gru-ode-bayes}
\begin{algorithmic}
   \STATE {\bfseries Input:} Initial state $\mathbf{h}_0$,
   \STATE \quad\quad observations $\mathbf{y}$, mask $\mathbf{m}$,
   \STATE \quad \quad observation times $\mathbf{t}$, final time $T$.
   \STATE Initialize $\mathrm{time} = 0, \mathrm{loss} = 0$, $\mathbf{h} = \mathbf{h}_0$.
   \FOR{$k=1$ {\bfseries to} $K$}
        \STATE \COMMENT{{\color{ForestGreen} ODE evolution to $\mathbf{t}[k]$}}
       \STATE $\mathbf{h} = \gruode(\mathbf{h}, \mathrm{time}, \mathbf{t}[k])$  
       \STATE $\mathrm{time} = t[k]$
       \STATE \COMMENT{{\color{ForestGreen} Pre-jump loss}}
       \STATE $\mathrm{loss} \mathrel{+}= \loss_{\mathrm{pre}}(\mathbf{y}[k], \mathbf{m}[k], \mathbf{h})$ 
       \STATE \COMMENT{{\color{ForestGreen} Update}}
       \STATE $\mathbf{h} = \grubayes(\mathbf{y}[k], \mathbf{m}[k], \mathbf{h})$  
       \STATE \COMMENT{{\color{ForestGreen} Post-jump loss}}
       \STATE $\mathrm{loss} \mathrel{+}= \lambda . \loss_{\mathrm{post}}(\mathbf{y}[k], \mathbf{m}[k], \mathbf{h})$ 
   \ENDFOR
    \STATE \COMMENT{{\color{ForestGreen} ODE evolution to $T$}}
   \STATE $\mathbf{h} = \gruode(\mathbf{h}, \mathbf{t}[N_K], T)$  
   \STATE {\bfseries return} $(\mathbf{h}, \mathrm{loss})$
\end{algorithmic}
\end{algorithm}
\end{minipage}
\end{wrapfigure}

With $p_{\mathrm{obs}}$, the PDF of $\mathbf{Y}(t)$ given the noisy observation (with noise vector $\bm{\epsilon}$), we first compute the analogue of the Bayesian update:
\begin{equation*}
    p_{\mathrm{Bayes},j} \propto  p_{\mathrm{pre}, j} \cdot p_{\mathrm{obs}, j}.
\end{equation*}

Let $p_{\mathrm{post}}$ denote the predicted distribution (from $\mathbf{h}_+$) after applying GRU-Bayes. We then define the post-jump loss as the KL-divergence between $ p_{\mathrm{Bayes}}$  and $p_{\mathrm{post}}$:
\begin{align*}
    \loss_{\mathrm{post}} = \sum_{j=1}^D m_j D_{KL}(p_{\mathrm{Bayes}, j} || p_{\mathrm{post}, j}).
\end{align*}
In this way, we force our model to learn to mimic a Bayesian update.


Similarly to the pre-jump loss, $\loss_{\mathrm{post}}$ is computed only for the observed dimensions. The total loss is then obtained by adding  both losses with a weighting parameter $\lambda$.

For binomial and Gaussian distributions, computing $\loss_{\mathrm{post}}$ can be done analytically. In the case of Gaussian distribution we can compute the Bayesian updated mean $\mu_{\mathrm{Bayes}}$ and variance $\sigma^2_{\mathrm{Bayes}}$ as
\begin{align*}
\mu_{\mathrm{Bayes}} &=
    \dfrac{\sigma^2_{\mathrm{obs}}}{\sigma^2_{\mathrm{pre}}+\sigma^2_{\mathrm{obs}}} \mu_{\mathrm{pre}} +
    \dfrac{\sigma^2_{\mathrm{pre}}}{\sigma^2_{\mathrm{pre}}+\sigma^2_{\mathrm{obs}}} \mu_{\mathrm{obs}}
\\
\sigma^2_{\mathrm{Bayes}} &= \frac{\sigma^2_{\mathrm{pre}} . \sigma^2_{\mathrm{obs}}}{\sigma^2_{\mathrm{pre}} + \sigma^2_{\mathrm{obs}}},
\end{align*}
where for readability we dropped the dimension sub-index.
In many real-world cases, the observation noise $\sigma^2_{\mathrm{obs}} \ll \sigma^2_{\mathrm{pre}}$, in which case $p_{\mathrm{Bayes}}$ is just the observation distribution: $\mu_{\mathrm{Bayes}} = \mu_{\mathrm{obs}}$ and $\sigma^2_{\mathrm{Bayes}} = \sigma^2_{\mathrm{obs}}$.

\subsection{Implementation}
The pseudocode of \method{} is depicted in Algorithm~\ref{alg:gru-ode-bayes}, where a forward pass is shown for a single time series~\footnote{Code is available in the following anonymous repository : \url{https://github.com/edebrouwer/gru_ode_bayes} }.  For mini-batching several time series we sort the observation times across all time series and for each unique time point $\mathbf{t}[k]$, we create a list of the time series that have observations. The main loop of the algorithm iterates over this set of unique time points. In the GRU-ODE step, we propagate all hidden states jointly. The GRU-Bayes update and the loss calculation are only executed on the time series that have observation at that particular time point. The complexity of our approach then scales linearly with the number of observations and quadratically with the dimension of the observations. When memory cost is a bottleneck, the gradient can be computed using the adjoint method, without backpropagating through the solver operations~\citep{neural_ode}.

\section{Related research}

Machine learning has a long history in time series modelling \citep{mitchell1999machine,gers2000learning,wang2006gaussian,chung2014empirical}. However, recent massive real-world data collection, such as electronic health records (EHR), increase the need for models capable of handling such complex data \citep{lee2017big}. As stated in the introduction, their \emph{sporadic} nature is the main difficulty. 

To address the nonconstant sampling, a popular approach is to recast observations into fixed duration time bins. However, this representation results in missing observation both in time and across features dimensions. This makes the direct usage of neural network architectures tricky. To overcome this issue, the main approach consists in some form of data imputation and jointly feeding the observation mask and times of observations to the recurrent network \citep{che2018recurrent,choi2016doctor,lipton2016directly,du2016recurrent,choi2016retain,BRITS}. This approach strongly relies on the assumption that the network will learn to process true and imputed samples differently. Despite some promising experimental results, there is no guarantee that it will do so. Some researchers have tried to alleviate this limitation by introducing more meaningful data representation for sporadic time series   \citep{rajkomar2018scalable,razavian2015temporal,ghassemi2015multivariate}, like tensors~\citep{de2018deep,simm2017macau}. 

Others have addressed the missing data problem with generative probabilistic models. Among those, (multitask) Gaussian processes (GP) are the most popular by far \citep{bonilla2008multi}. They have been used for smart imputation before a RNN or CNN architecture \citep{futoma2017learning, moor2019temporal}, for modelling a hidden process in joint models \citep{soleimani2018scalable}, or to derive informative representations of time series \citep{ghassemi2015multivariate}. GPs have also been used for direct forecasting \citep{cheng2017sparse}. However, they usually suffer from high uncertainty outside the observation support, are computationally intensive \citep{quinonero2005unifying}, and learning the optimal kernel is tricky. Neural Processes, a neural version of GPs, have also been introduced by \citet{garnelo2018neural}. In contrast with our work that focuses on continuous-time \emph{real-valued} time series, continuous time modelling of time-to-events has been addressed with point processes \citep{mei2017neural} and continuous time Bayesian networks \citep{nodelman2002continuous}. Yet, our continuous modelling of the latent process allows us to straightforwardly model a continuous intensity function and thus handle both real-valued and event type of data. This extension was left for future work.

Most recently, the seminal work of \citet{neural_ode} suggested a continuous version of neural networks that overcomes the limits imposed by discrete-time recurrent neural networks. Coupled with a variational auto-encoder architecture \citep{kingma2013auto}, it proposed a natural way of generating irregularly sampled data. However, it transferred the difficult task of processing sporadic data to the encoder, which is a discrete-time RNN. In a work submitted concomitantly to ours \citep{rubanova2019latent}, the authors proposed a convincing new VAE architecture that uses a Neural-ODE architecture for both encoding and decoding the data.

Related auto-encoder approaches with sequential latents operating in discrete time have also been proposed \citep{krishnan2015deep,krishnan2017structured}. These models rely on classical RNN architectures in their inference networks, hence not addressing the sporadic nature of the data. What is more, if they have been shown useful for smoothing and counterfactual inference, their formulation is less suited for forecasting. Our method also has connections to the Extended Kalman Filter (EKF) that models the dynamics of the distribution of processes in continuous time. However, the practical applicability of the EKF is limited because of the linearization of the state update and the difficulties involved in identifying its parameters. Importantly, the ability of the GRU to learn long-term dependencies is a significant advantage. 

Finally, other works have investigated the relationship between deep neural networks and partial differential equations. An interesting line of research has focused on deriving better deep architectures motivated by the stability of the corresponding patial differential equations (PDE) \citep{haber2017stable,chang2019antisymmetricrnn}. Despite their PDE motivation, those approaches eventually designed new \emph{discrete} architectures and didn't explore the application on continuous inputs and time.

\section{Application to synthetic SDEs}
\label{sec:SyntheticSDEs}
Figure \ref{fig:comparison} illustrates the capabilities of our approach compared to NeuralODE-VAE on data generated from a process driven by a multivariate Ornstein-Uhlenbeck (OU) SDE with random parameters. Compared to NeuralODE-VAE, which retrieves the average dynamics of the samples, our approach detects the correlation between both features and updates its predictions more finely as new observations arrive. In particular, note that \method{} updates its prediction and confidence on a feature even when only the other one is observed, taking advantage from the fact that they are correlated. This can be seen on the left pane of Figure~\ref{fig:comparison} where at time $t=3$, Dimension~1 (blue) is updated because of the observation of Dimension~2 (green). 

By directly feeding sporadic inputs into the ODE, \method{} sequentially \emph{filters} the hidden state and thus estimates the PDF of the future observations. This is the core strength of the proposed method, allowing it to perform long-term predictions.

In Appendix~\ref{app:OU_SDE}, we further show that our model can exactly represent the dynamics of multivariate OU process with random variables. Our model can also handle nonlinear SDEs as shown in Appendix~\ref{app:BXLator} where we present an example inspired by the Brusselator~\citep{prigogine1982being}, a chaotic ODE.


\section{Empirical evaluation}
\label{sec:real_world}

We evaluated our model on two data sets from different application areas: healthcare and climate forecasting. In both applications, we assume the data consists of noisy observations from an underlying unobserved latent process as in Eq.~\ref{eq:SDE}. We focused on the general task of forecasting the time series at future time points. Models are trained to minimize negative log-likelihood.

\subsection{Baselines}

We used a comprehensive set of state-of-the-art baselines to compare the performance of our method. All models use the same hidden size representation and comparable number of parameters.

\textbf{NeuralODE-VAE} \citep{neural_ode}. We model the time derivative of the hidden representation as a 2-layer MLP. To take missingness across features into account, we add a mechanism to feed an observation mask.

\textbf{Imputation Methods.} We implemented two imputation methods as described in \citet{che2018recurrent}: \emph{GRU-Simple} and \emph{GRU-D}. 

\textbf{Sequential VAEs} \citep{krishnan2015deep,krishnan2017structured}. We extended the deep Kalman filter architecture by feeding an observation mask and updating the loss function accordingly.

\textbf{T-LSTM} \citep{baytas2017patient}. We reused the proposed time-aware LSTM cell to design a forecasting RNN with observation mask.

\subsection{Electronic health records}
Electronic Health Records (EHR) analysis is crucial to achieve data-driven personalized medicine~\citep{lee2017big,goldstein2017opportunities,esteva2019guide}. However, efficient modeling of this type of data remains challenging. Indeed, it consists of \emph{sporadically} observed longitudinal data with the extra hurdle that there is no standard way to align patients trajectories (\emph{e.g.}, at hospital admission, patients might be in very different state of progression of their condition). Those difficulties make EHR analysis well suited for GRU-ODE-Bayes.

We use the publicly available MIMIC-III clinical database \citep{johnson2016mimic}, which contains EHR for more than 60,000 critical care patients. We select a subset of 21,250 patients with sufficient observations and extract 96 different longitudinal real-valued measurements over a period of 48 hours after patient admission. We refer the reader to Appendix~\ref{app:mimic3} for further details on the cohort selection. We focus on the predictions of the next 3 measurements after a 36-hour observation window. 

\subsection{Climate forecast}

From short-term weather forecast to long-range prediction or assessment of systemic changes, such as global warming, climatic data has always been a popular application for time-series analysis. This data is often considered to be regularly sampled over long periods of time, which facilitates their statistical analysis. Yet, this assumption does not usually hold in practice. Missing data are a problem that is repeatedly encountered in climate research because of, among others, measurement errors, sensor failure, or faulty data acquisition. The actual data is then sporadic and researchers usually resort to imputation before statistical analysis  \citep{junninen2004methods,schneider2001analysis}.

We use the publicly available United State Historical Climatology Network (USHCN) daily data set~\citep{ushcn}, which contains measurements of 5 climate variables (daily temperatures, precipitation, and snow) over 150 years for 1,218 meteorological stations scattered over the United States. We selected a subset of 1,114 stations and an observation window of 4 years (between 1996 and 2000). To make the time series sporadic, we subsample the data such that each station has an average of around 60 observations over those 4 years. Appendix~\ref{app:ushcn} contains additional details regarding this procedure. The task is then to predict the next 3 measurements after the first 3 years of observation.

\subsection{Results}
We report the performance using 5-fold cross-validation. Hyperparameters (dropout and weight decay) are chosen using an inner holdout validation set (20\%) and performance are assessed on a left-out test set (10\%). Those folds are reused for each model we evaluated for sake of reproducibility and fair comparison (More details in Appendix \ref{app:hyper-params}). Performance metrics for both tasks (NegLL and MSE) are reported in Table~\ref{table:forecast}. \method~handles the sporadic data more naturally and can more finely model the dynamics and correlations between the observed features, which results in higher performance than other methods for both data sets. In particular, \method{} unequivocally outperforms all other methods on both data sets.

\begin{table}[htb]
\caption{Forecasting results.}
\label{table:forecast}
\vskip 0.15in
\begin{center}
\begin{small}
\begin{sc}
\begin{tabular}{lrr|rr}
\toprule
& \multicolumn{2}{c}{USHCN-Daily} & \multicolumn{2}{c}{MIMIC-III} \\
\midrule
Model & MSE & NegLL & MSE & NegLL \\
\midrule
NeuralODE-VAE   &       $0.96 \pm 0.11 $ & $1.46 \pm 0.10$ & $0.89 \pm 0.01 $ & $1.35 \pm 0.01$ \\
NeuralODE-VAE-Mask & $0.83 \pm 0.10$ & $1.36 \pm 0.05$ & $0.89 \pm 0.01$ & $1.36 \pm 0.01$  \\
Sequential VAE & $0.83 \pm 0.07$ & $ 1.37 \pm  0.06$ & $0.92 \pm 0.09$ & $1.39 \pm 0.07$\\
GRU-Simple  &  $0.75 \pm 0.12$ & $1.23 \pm 0.10$ &  $0.82 \pm 0.05$ & $1.21 \pm 0.04$\\
GRU-D  &  $0.53 \pm 0.06$ & $0.99 \pm 0.07$  &  $0.79 \pm 0.06$ & $1.16 \pm 0.05$\\
T-LSTM & $0.59 \pm 0.11$ & $1.67 \pm 0.50$ & $0.62 \pm 0.05$ & $1.02 \pm 0.02$\\
\hline
GRU-ODE-Bayes     &  $\mathbf{0.43} \pm 0.07$ & $\mathbf{0.84} \pm 0.11$   &  $\mathbf{0.48} \pm 0.01$ & $\mathbf{0.83} \pm 0.04$\\

\bottomrule
\end{tabular}
\end{sc}
\end{small}
\end{center}
\vskip -0.1in
\end{table}

\subsection{Impact of continuity prior}
\label{sec:continuity}
To illustrate the capabilities of the derived GRU-ODE cell presented in Section \ref{sec:gru-ode}, we consider the case of time series forecasting with low sample size. In the realm of EHR prediction, this could be framed as a \emph{rare disease} setup, where data is available for few patients only. In this case of scarce number of samples, the continuity prior embedded in GRU-ODE is crucially important as it provides important prior information about the underlying process.

To highlight the importance of the GRU-ODE cell, we compare two versions of our model : the classical \method{} and one where the GRU-ODE cell is replaced by a discretized autonomous GRU. We call the latter \emph{GRU-Discretized-Bayes}. Table \ref{table:gru-disc} shows the results for MIMIC-III with varying number of patients in the training set. While our discretized version matches the continuous one on the full data set, GRU-ODE cell achieves higher accuracy when the number of samples is low, highlighting the importance of the continuity prior. Log-likelihood results are given in Appendix~\ref{app:small-sample}.

\begin{savenotes}
\begin{table}[htb]
\caption{Comparison between GRU-ODE and discretized version in the small-sample regime (MSE).}
\label{table:gru-disc}
\vskip 0.15in
\begin{center}
\begin{small}
\begin{sc}
\begin{tabular}{lccc}
\toprule
Model & \multicolumn{1}{c}{1,000 patients} & \multicolumn{1}{c}{2,000 patients} & \multicolumn{1}{c}{Full}  \\
\midrule


NeuralODE-VAE-Mask       & $0.94 \pm 0.01$  & $0.94 \pm 0.01$  & $0.89 \pm 0.01$  \\ 
GRU-Discretized-Bayes &  $0.87 \pm 0.02$  & $0.77 \pm 0.02$  & $\mathbf{0.46} \pm 0.05$  \\
GRU-ODE-Bayes     &  $\mathbf{0.77} \pm 0.01$ & $\mathbf{0.72} \pm 0.01$ &  $\mathbf{0.48}$\daggerfootnote{Statistically not different from best (p-value $>0.6$ with paired t-test).} $ \pm 0.01$  \\

\bottomrule
\end{tabular}
\end{sc}
\end{small}
\end{center}
\vskip -0.1in
\end{table}
\end{savenotes}

\section{Conclusion and future work}

We proposed a model combining two novel techniques, GRU-ODE and GRU-Bayes, which allows feeding sporadic observations into a  continuous ODE dynamics describing the evolution of the probability distribution of the data. Additionally, we showed that this filtering approach enjoys attractive representation capabilities. Finally, we demonstrated the value of \method{} on both synthetic and real-world data. 
Moreover, while a discretized version of our model performed well on the full MIMIC-III data set, the continuity prior of our ODE formulation proves particularly important in the small-sample regime, which is particularly relevant for real-world clinical data where many data sets remain relatively modest in size.

In this work, we focused on time-series data with Gaussian observations. However, \method{} can also be extended to binomial and multinomial observations since the respective NegLL and KL-divergence are analytically tractable. This allows the modeling of sporadic observations of both discrete and continuous variables. 
\section*{Acknowledgements}

Edward De Brouwer is funded by a FWO-SB grant. Yves Moreau is funded by (1) Research Council KU Leuven: C14/18/092 SymBioSys3; CELSA-HIDUCTION, (2) Innovative Medicines Initiative: MELLODY, (3) Flemish Government (ELIXIR Belgium, IWT, FWO 06260) and (4) Impulsfonds AI: VR 2019 2203 DOC.0318/1QUATER Kenniscentrum Data en Maatschappij. Computational resources and services used in this work were provided by the VSC (Flemish Supercomputer Center), funded by the Research Foundation - Flanders (FWO) and the Flemish Government – department EWI. We also gratefully acknowledge the support of NVIDIA Corporation with the donation of the Titan Xp GPU used for this research.

\vfill

\bibliography{gruode}
\bibliographystyle{icml2019}

\newpage

\appendix

\section{Full formulation of the GRU-ODE cell}
\label{app:gru-ode-full}
The full ODE equation for GRU-ODE is the following:

\begin{align*}
  \dfrac{d \mathbf{h}(t)}{dt} =  (1 - \mathbf{z}(t)) \odot (\mathbf{g}(t) - \mathbf{h}(t)),
\end{align*}

with

\begin{align*}
  \mathbf{r}(t) &= \sigma(W_r \mathbf{x}(t) + U_r \mathbf{h}(t)+\mathbf{b}_r)  \\
  \mathbf{z}(t) &= \sigma(W_z \mathbf{x}(t) + U_z \mathbf{h}(t)+\mathbf{b}_z) \\
  \mathbf{g}(t) &= \tanh(W_h  \mathbf{x}(t) + U_h (\mathbf{r}(t) \odot \mathbf{h}(t)) +  \mathbf{b}_h).
  \end{align*}
  
Matrices $W \in \mathbb{R}^{H\times D}$, $B \in \mathbb{R}^{H\times H}$ and bias vectors $\mathbf{b} \in \mathbb{R}^{H}$ are the parameters of the cell. $H$ and $D$ are the dimension of the hidden process and of the inputs respectively.

\section{Lipschitz continuity of GRU-ODE}
\label{app:continuity}
As $\mathbf{h}$ is differentiable and continous on $t$, we know from the mean value theorem that for any $t_a, t_b \in t$, there exists $t^* \in (t_a, t_b)$ such that 
$$ \mathbf{h}(t_b)-\mathbf{h}(t_a) = \frac{d\mathbf{h}}{dt}\mid_{t^*} (t_b-t_a).$$
Taking the euclidean norm of the previous expression, we find 
$$\mid \mathbf{h}(t_b)-\mathbf{h}(t_a) \mid = \mid \frac{d\mathbf{h}}{dt}\mid_{t^*}   (t_b-t_a) \mid . $$
Furthermore, we showed that $\mathbf{h}$ is bounded on $[-1,1]$. Hence, because of the bounded functions appearing in the ODE (sigmoids and hyperbolic tangents), the derivative of $\mathbf{h}$ is itself bounded by $[-2,2]$. We conclude that $\mathbf{h}(t)$ is Lipschitz continuous with constant $K=2$.

\section{Comparison of numerical integration methods}
\label{app:dopri}
We implemented three numerical integration methods, among which the classical Euler method and the Dormand-Prince method (DOPRI). DOPRI is a popular adaptive step size numerical integration method relying on 2 Runge-Kutta solvers of order 4 and 5. The advantage of adaptive step size methods is that they can tune automatically the number of steps to integrate the ODE until the desired point.

\begin{figure}[H]
\vskip 0.0in
\begin{center}
\centerline{\includegraphics[width=10cm]{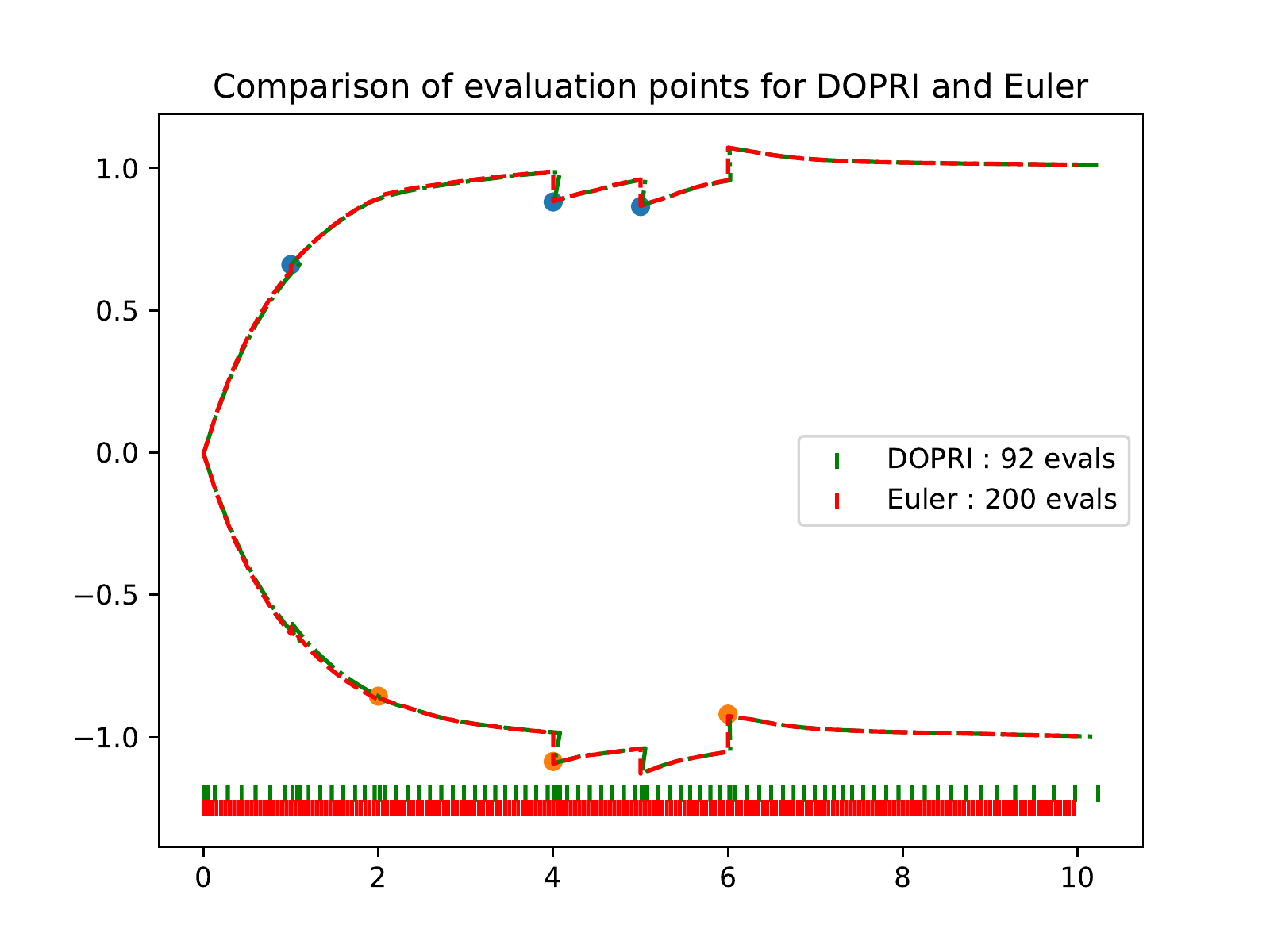}}
\caption{Comparison of Euler and DOPRI numerical integration methods for same inputs and same ODE. Colored ticks on the $x$ axis represent the evaluation time for each method. Dotted lines show the evolution of the estimated mean distribution of the observations while the dots stand for the observations fed to the model.}
\label{fig:RK-doubleOU}
\end{center}
\vskip -0.2in
\end{figure}

Figure~\ref{fig:RK-doubleOU} illustrates the number of steps taken by both solvers when given the same data and same ODE. We observe that using an adaptive step size results in half as many time steps. More steps are taken near the observations and as the underlying process becomes smoother, the step size increase, as observed on the right side of the figure. However, each time step requires significantly fewer computations for Euler than for DOPRI, so that Euler's method appears more than competitive on the data and simulations we have considered so far. Nevertheless, DOPRI might still be preferred as default method because of its better numerical stability.

\section{Mapping to deal with missingness across features}
\label{app:fprep}
The preprocessing step $f_{\mathrm{prep}}$ for GRU-Bayes takes in the hidden state $\mathbf{h}$ and computes the parameters for the observation PDFs $\theta = f_{\mathrm{obs}}(\mathbf{h}(t))$.
In the case of a Gaussian, $\theta_d$  contains the means and log-variances for dimension $d$ of $\mathbf{Y}(t)$.
Then, we create a vector $\mathbf{q}_d$ that concatenates $\theta_d$ with  the observed value $\mathbf{y}[k]_d$ and the normalized error term, which for the Gaussian case is $(\mathbf{y}[k] _d- \mu_d) / \sigma_d$, where $\mu_d$ and $\sigma_d$ are the mean and standard deviation derived from $\theta_d$. We then multiply the vectors $\mathbf{q}_d$ by a dimension-specific weight matrix $\mathbf{W}_d$ and apply a ReLU nonlinearity. Next, we zero all results that did not have an observation (by multiplying them with mask $m_{d}$). Finally, the concatenation of the results is fed into the GRU unit of GRU-Bayes.

\section{Observation model mapping}
\label{app:f_obs}

The mapping from hidden $\mathbf{h}$ to the parameters of the distribution $\mu_{\mathbf{Y}(t)}$ and $\log (\Sigma_{\mathbf{Y}(t)})$. For this purpose we use a classical multi-layer perceptron architecture with a 25 dimensional hidden layer. Note that me map to the log of the variance in order to keep it positive.

\section{GRU-ODE-Bayes-seq}
\label{app:gru-seq}
On top of the architecture described in the main bulk of this paper, we also propose a variant which process the sporadic inputs \emph{sequentially}. In other words, GRU-Bayes will update its prediction on the hidden $\mathbf{h}$ for one input dimension after the other rather than jointly. We call this approach GRU-ODE-Bayes-seq.

In this sequential approach for GRU-Bayes, we process one-by-one all dimensions of $\mathbf{y}[k]$  that were \emph{observed} at time $t[k]$ by first applying the preprocessing to each and then sending them to the GRU unit.
The preprocessing steps are the same as in the nonsequential scheme (Appendix~\ref{app:fprep}) but without concatenation at the end because only one dimension is processed at a time.
Note that the preprocessing of dimensions cannot be done in parallel as the hidden state $\mathbf{h}$ changes after each dimension is processed, which affects the computed $\theta_d$  and thus the resulting vector $\mathbf{q}_d$.

\section{Minimal GRU-ODE}
\label{app:gru-minimal}
Following the same reasoning as for the full GRU cell, we also derived the minimal GRU-ODE cell, based on the minimal GRU cell. The minimal GRU writes :

\begin{align*}
  \mathbf{f}_t &= \sigma(W_f \mathbf{x}_t + U_f \mathbf{h}_{t-1}+\mathbf{b}_f) \nonumber \\
  \mathbf{h}_t &= \mathbf{f}_t \odot \mathbf{h}_{t-1} + (1-\mathbf{f}_t) \odot \sigma(W_h \mathbf{x}_t + U_h( \mathbf{h}_{t-1}\odot \mathbf{f}_t )+\mathbf{b}_h) \nonumber \\
\end{align*}

This can be rewritten as the following difference equation :

\begin{equation*}
	\Delta \mathbf{h}_t = (1-\mathbf{f}_t) \odot (\sigma(W_h \mathbf{x}_t + U_h( \mathbf{h}_{t-1}\odot \mathbf{f}_t )+\mathbf{b}_h) - \mathbf{h}_{t-1})
\end{equation*}

Which leads to the corresponding ODE :

\begin{align*}
  \dfrac{d \mathbf{h}(t)}{dt} =  (1-\mathbf{f}(t)) \odot (\sigma(W_h \mathbf{x}(t) + U_h( \mathbf{h}(t)\odot \mathbf{f}(t) )+\mathbf{b}_h) - \mathbf{h}(t))
\end{align*}

\section{Ablation study of GRU-Bayes}
\label{app:ablation_bayes}
In order to demonstrate the fitness of the GRU-Bayes module for our architecture, we ran an ablation study where we replaced the GRU-Bayes with a 2 layers multi-layer perceptron. We used a \emph{tanh} activation function for the hidden units and a linear activation for the output layer. We evaluate the performance of this modified architecture on the MIMIC dataset for the forecasting task. The results are presented in table~\ref{table:ablationGRUBayes}. The proposed architecture outperforms a simple MLP module, due to the properties described in sections \ref{subsec:gru_ode_properties} and \ref{subsec:gru_bayes}.

\begin{table*}[h]
\caption{NegLL and MSE results for proposed GRU-Bayes module and replaced with MLP.}
\label{table:ablationGRUBayes}
\vskip 0.15in
\begin{center}
\begin{small}
\begin{sc}
\begin{tabular}{lcc}
\toprule
Model & MSE & NegLL  \\
\midrule
GRU-Bayes original & $0.48 \pm 0.01$ & $0.83 \pm 0.04$  \\
GRU-Bayes MLP &   $0.54 \pm 0.05$ & $1.05 \pm 0.02$  \\

\bottomrule
\end{tabular}
\end{sc}
\end{small}
\end{center}
\vskip -0.1in
\end{table*}

\section{Application to Ornstein-Uhlenbeck SDEs}
\label{app:OU_SDE}
We demonstrate the capabilities of our approach on data generated from a process driven by an SDE as in Eq.~\ref{eq:SDE}. In particular, we focus on extensions of the multidimensional correlated Ornstein-Uhlenbeck (OU) process with varying parameters. For a particular sample $i$, the dynamics is given by the following SDE:
\begin{align}
d \mathbf{Y}_i(t) = \theta_i(\mathbf{r}_i-\mathbf{Y}_i(t))dt + \sigma_i d\mathbf{W}(t),
\label{eq:OU}
\end{align}
where $\mathbf{W}(t)$ is a \(D\)-dimensional correlated Wiener process, $\mathbf{r}_i$ is the vector of targets, and $\theta_i$ is the reverting strength constant. For simplicity, we consider $\theta_i$ and $\sigma_i$ parameters as scalars. Each sample $\mathbf{y}_i$ is then obtained via the realization of process~\eqref{eq:OU} with sample-specific parameters.

\subsection{Representation capabilities}
\label{sec:representation-capabilities}
We now show that our model exactly captures the dynamics of the distribution of $\mathbf{Y}(t)$ as defined in Eq.~\ref{eq:OU}. The evolution of the PDF of a diffusion process is given by the corresponding Fokker-Planck equation. For the OU process, this PDF is Gaussian with time-dependent mean and covariance. Conditioned on a previous observation at time $t^*$, this gives
\begin{align*}
    \mathbf{Y}_i(t) \mid \mathbf{Y}_i(t^*) &\sim \mathcal{N}(\mu_{\mathbf{Y}}(t,t^*),\sigma_{\mathbf{Y}}^2(t,t^*)), \\
    \mu_{\mathbf{Y}}(t,t^*) &= \mathbf{r}_i + (\mathbf{Y}_i(t^*)-\mathbf{r}_i)\exp{(-\theta_i (t - t^*))}, \\
    \sigma_{\mathbf{Y}}^2(t,t^*) &= \frac{\sigma^2_i}{2\theta_i} (1-\exp(-2\theta_i (t-t^*))).
\end{align*}
Correlation of $\mathbf{Y}(t)$ is constant and equal to $\rho$, the correlation of the Wiener processes. The dynamics of the mean and variance parameters can be better expressed in the following ODE form:
\begin{equation}
\begin{split}
\dfrac{d \mu_{\mathbf{Y}}(t,t^*)}{dt} &= -\theta_i (\mu_{\mathbf{Y}}(t,t^*)-\mathbf{r}_i) \\
\dfrac{d \sigma_{\mathbf{Y}}^2(t,t^*)}{dt} &= -2\theta_i \left(\sigma_{\mathbf{Y}}^2(t,t^*) - \frac{\sigma_i^2}{2 \theta_i}\right)
\label{eq:OU_params}
\end{split}
\end{equation}
With initial conditions $\mu_{\mathbf{Y}}(0,t^*)= \mathbf{Y}(t^*)$ and $\sigma_{\mathbf{Y}}^2(0,t^*)=0$. We next investigate how specific versions of this ODE can be represented by our \method{}.

\subsubsection{Standard Ornstein-Uhlenbeck process}
In standard OU, the parameters $\mathbf{r}_i$, $\sigma_i$, and $\theta_i$ are fixed and identical for all samples. The ODE~\eqref{eq:OU_params} is linear and can then be represented directly with GRU-ODE by storing $\mu_{\mathbf{Y}}(t,t^*)$ and $\sigma_{\mathbf{Y}}^2(t,t^*)$ in the hidden state $\mathbf{h}(t)$ and matching the Equations~\eqref{eq:gru-ode-update} and \eqref{eq:OU_params}. The OU parameters $\mathbf{r}_i$, $\sigma_i$ and $\theta_i$ are learned and encoded in the weights of GRU-ODE. GRU-Bayes then updates the hidden state and stores $\mu_{\mathbf{Y}}(t,t^*)$ and $\sigma_{\mathbf{Y}}^2(t,t^*)$.

\subsubsection{Generalized Ornstein-Uhlenbeck processes}
When parameters are allowed to vary over samples, these have to be encoded in the hidden state of GRU-ODE-Bayes, rather than in the fixed weights. For $\mathbf{r}_i$ and $\sigma_i$, GRU-Bayes  computes and stores their current estimates as the observations arrive. This is based on previous hidden and current observation as in Eq.~\ref{eq:GRU-Bayes}.
The GRU-ODE module then simply has to keep these estimates unchanged between observations: $$\frac{d \mathbf{r}_i(t)}{dt}=\frac{d \sigma_i(t)}{dt} = 0.$$ This can be easily done by switching off the update gate (\emph{i.e.}, setting $\mathbf{z}(t)$ to 1 for these dimensions).
These hidden states can then be used to output the mean and variance in Eq.~\ref{eq:OU_params}, thus enabling the model to represent generalized Ornstein-Uhlenbeck processes with sample-dependent $\mathbf{r}_i$ and $\sigma_i$.

Perfect representation for sample dependent $\theta_i$ requires the multiplication of inputs in Eq.~\ref{eq:OU_params}, which GRU-ODE is not able to perform exactly but should be able to approximate reasonably well. If an exact representation is required, the addition of a bilinear layer is sufficient.
    
Furthermore, the same reasoning applies when parameters are also allowed to change over time within the same sample. GRU-Bayes is again able to update the hidden vector with the new estimates.

\subsubsection{Non-aligned time series}
Our approach can also handle samples that would be dephased in time
(\emph{i.e}, the observation windows are not aligned on an intrinsic time scale). 
Longitudinal patient data recorded at different stages of the disease for each patient is one example, developed in Section~\ref{sec:real_world}. This setting is naturally handled by the GRU-Bayes module.


\subsection{Case Study: 2D Ornstein-Uhlenbeck Process}
\label{sec:2d-ou-experiments}
\subsubsection{Setup}
We evaluate our model on a 2-dimensional OU process with correlated Brownian motion as defined in Eq.~\ref{eq:OU}. For best illustration of its capabilities, we consider the three following cases. 

In the first setting, $\mathbf{r}_i$ varies across samples as $\mathbf{r}_i^1 \sim \mathcal{U}(0.5,1.5) $ and $\mathbf{r}_i^2 \sim \mathcal{U}(-1.5,-0.5)$. The correlation between the Wiener processes $\rho$ is set to $0.99$. We also set $\sigma=0.1$ and $\theta=1$. The second case, which we call \emph{random lag} is similar to the first one but adds an extra uniformly distributed random lag to each sample. Samples are then time shifted by some $\Delta_T \sim \mathcal{U}(0,0.5)$. The third setting is identical to the first but with $\rho=0$ (\emph{i.e.}, both dimensions are independent and no information is shared between them).

\begin{figure}[hbt]
\centering
\centerline{\includegraphics[width=8cm]{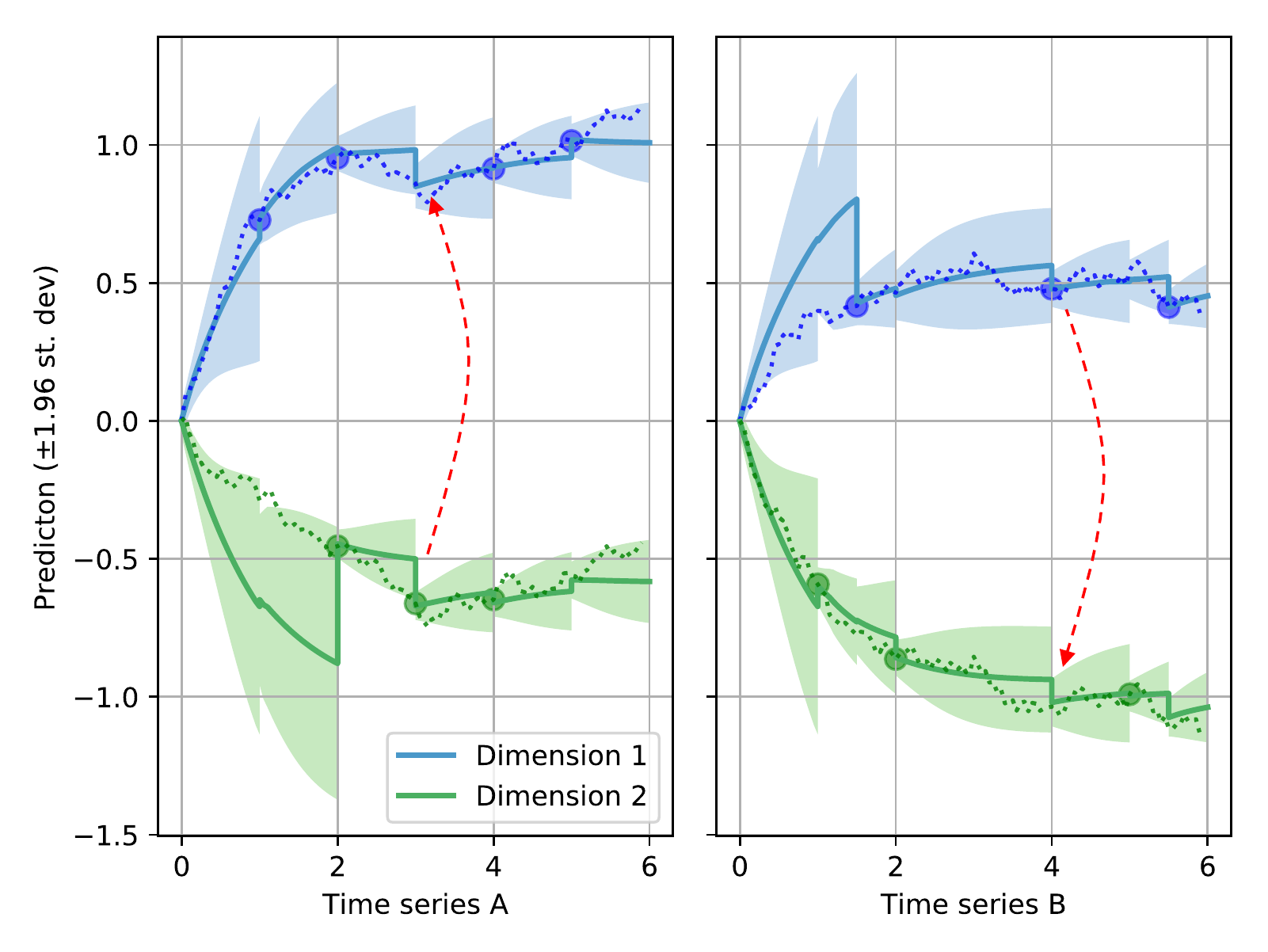}}
\caption{
Example of predictions (with shaded confidence intervals) given by \method{} for two samples of a correlated 2-dimensional stochastic process  (dotted line) with unknown parameters. Dots show the observations. Only a few observations are required for the model to infer the parameters. Additionally, \method{} learns the correlation between the dimensions resulting in updates of nonobserved variables (red dashed arrow). 
}
\label{fig:setup-figure}
\end{figure}

We evaluate all methods and settings on the forecast of samples after time $t=4$. The training set contains 10,000 samples with an average of 20 observations scattered over a 10-second time interval. Models are trained with a negative log-likelihood objective function, but mean square errors (MSE) are also reported. We compare our methods to NeuralODE-VAE \citep{neural_ode}. Additionally, we consider an extended version of this model where we also feed the observation mask, called NeuralODE-VAE-Mask.


\begin{table*}[t]
\caption{NegLL and MSE results for 2-dimensional general Ornstein-Uhlenbeck process.}
\label{table:double_ou-lik}
\vskip 0.15in
\begin{center}
\begin{small}
\begin{sc}
\begin{tabular}{lccc|cccr}
& \multicolumn{3}{c}{Negative Log-Likelihood} & \multicolumn{3}{c}{MSE}  \\
\toprule
Model & Random $r$ & Random Lag &  $\rho=0$ & Random r & Random Lag & $\rho=0$ \\
\midrule
NeuralODE-VAE-Mask & $0.222$ & $0.223$ &   $0.204$ & $0.081$ & $0.069$ &    $0.081$ \\
NeuralODE-VAE &   $0.183$ & $0.230$  &  $0.201$ & $0.085$ & $0.119$   & $0.113$ \\
\hline
GRU-ODE-Bayes                 &  $-1.260$ & $-1.234$ &  $-1.187$  & $0.005$ &  $0.005$ &    $0.006$\\
GRU-ODE-Bayes-minimal                 &  $-1.257$ & $-1.226$ & $-1.188$ & $0.005$ & $0.006$&    $0.006$ \\
GRU-ODE-Bayes-seq           &  $-1.266$ & $-1.225$ & $-1.191$  & $0.005$ & $0.005$ &    $0.006$  \\
GRU-ODE-Bayes-seq-minimal  &  $-1.260$ & $-1.225$ & $-1.189$ & $0.005$ & $0.005$  &  $0.006$ \\

\bottomrule
\end{tabular}
\end{sc}
\end{small}
\end{center}
\vskip -0.1in
\end{table*}

\subsubsection{Empirical evaluation}
Figure~\ref{fig:comparison} shows a comparison of predictions between NeuralODE-VAE and \method{} for the same sample issued from the random $\mathbf{r}_i$ setting. Compared to NeuralODE-VAE, which retrieves the average dynamics of the sample, our approach detects the correlation between both features and updates its predictions more finely as the observations arrive. In particular, note that \method{} updates its prediction and confidence on a feature even when only the other one is observed, taking advantage from the fact that they are correlated. This can be seen on the left pane of Figure~\ref{fig:comparison} where at time $t=3$ Dimension~1 (blue) is updated because of the observation of Dimension~2 (green). 

By directly feeding sporadic inputs into the ODE, \method{} sequentially \emph{filters} the hidden state and thus estimates the PDF of the future observations. This is the core strength of the proposed method, allowing it to perform long-term predictions. 
In contrast, NeuralODE-VAE first stores the whole dynamics in a single vector and later maps it to the dynamics of the time series (illustrated in Figure~\ref{fig:comparison}).

This analysis is confirmed by the performance results presented in Table~\ref{table:double_ou-lik}. Our approach performs better on all setups for both NegLL and MSE. What is more, the method deals correctly with \emph{lags} (\emph{i.e.}, the second setup) as it results in only marginal degradation of NegLL and MSE. When there is no correlation between both dimensions (\emph{i.e.,} $\rho=0$), the observation of one dimension contains no information on the other and this results in lower performance.

\begin{figure}[bht]
\vskip 0.0in
\begin{center}
\centerline{\includegraphics[width=10cm]{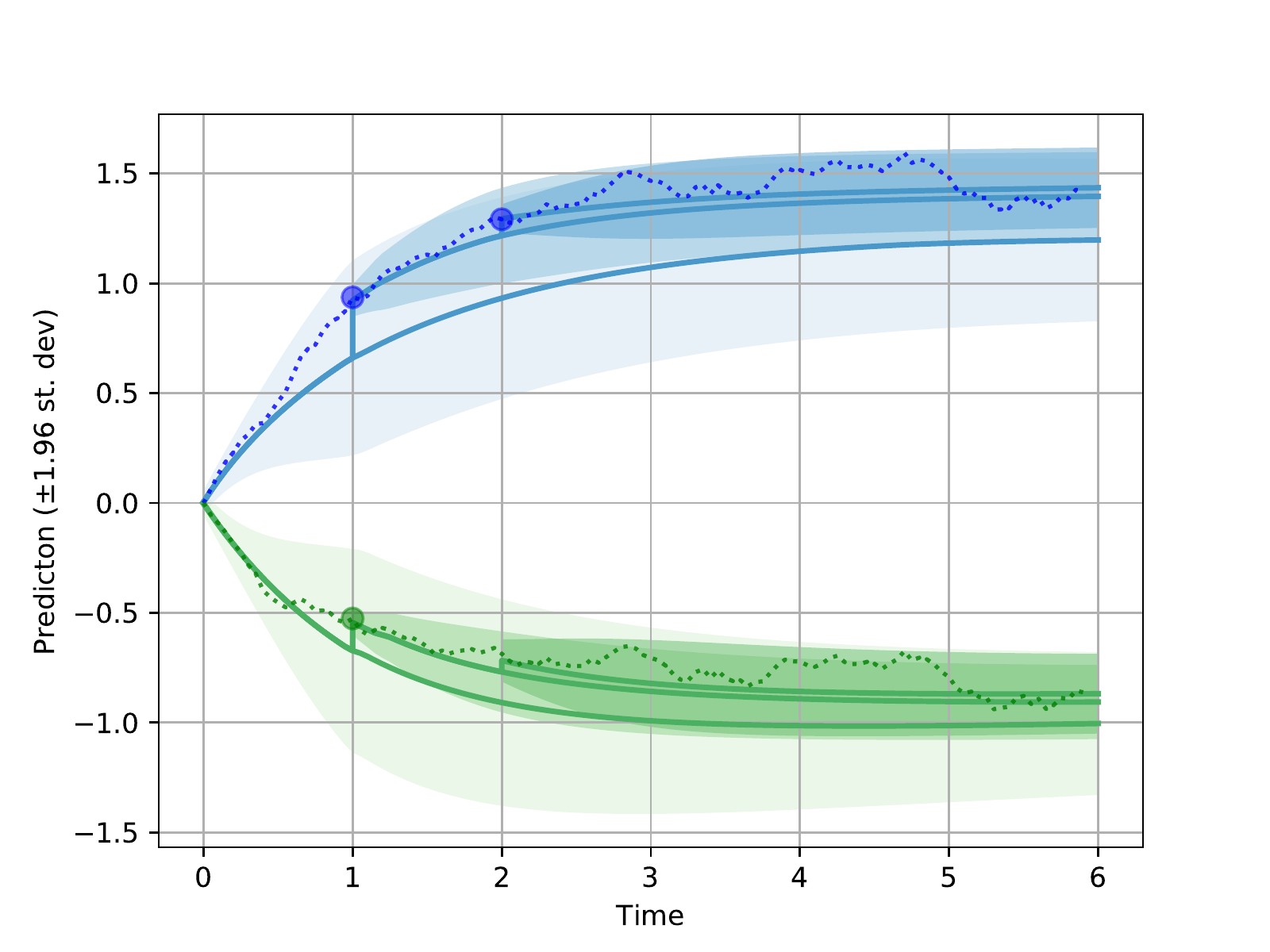}}
\caption{GRU-ODE-Bayes updating its prediction trajectory with every new observation for the random $\mathbf{r}_i$ setup. Shaded regions are propagated confidence intervals conditioned on previous observations. }
\label{fig:matriochka}
\end{center}
\vskip -0.2in
\end{figure}

Figure \ref{fig:matriochka} illustrates how \method{} updates its prediction and confidence as more and more observations are processed. This example is for the first setup (randomized $\mathbf{r}_i$).
Initially, the predictions have large confidence intervals and reflect the general statistics of the training data.
Then, observations gradually reduce the variance estimate as the model refines its predictions of the parameter $\mathbf{r}_i$. 
As more data is processed, the predictions converge to the asymptotic distribution of the underlying process.

\section{Application to synthetic nonlinear SDE: the Brusselator}
\label{app:BXLator}

On top of the extended multivariate OU process, we also studied a nonlinear SDE. We derived it from the Brusselator ODE, which was proposed by Ilya Prigogine to model autocatalytic reactions \citep{prigogine1982being}. It is a 2-dimensional process characterized by the following equations:

\begin{align*}
    \frac{dx}{dt} &= 1 + (b+1)x + ax^2y \\
    \frac{dy}{dt} &= bx -ax^2y 
\end{align*}

Where $x$ and $y$ stand for the two dimensions of the process and $a$ and $b$ are parameters of the ODE. This system becomes unstable when $b>1+a$. We add a stochastic component to this process to make it the following SDE, which we will model:

\begin{align}
\begin{split}
    \frac{dx}{dt} &= 1+(b+1)x + ax^2y + \sigma dW_1(t) \\
    \frac{dy}{dt} &= bx -ax^2y  + \sigma dW_2(t)
\end{split}
    \label{eq:BXLator}
\end{align}

\begin{figure}[hbt]
\centering
\centerline{\includegraphics[width=15cm]{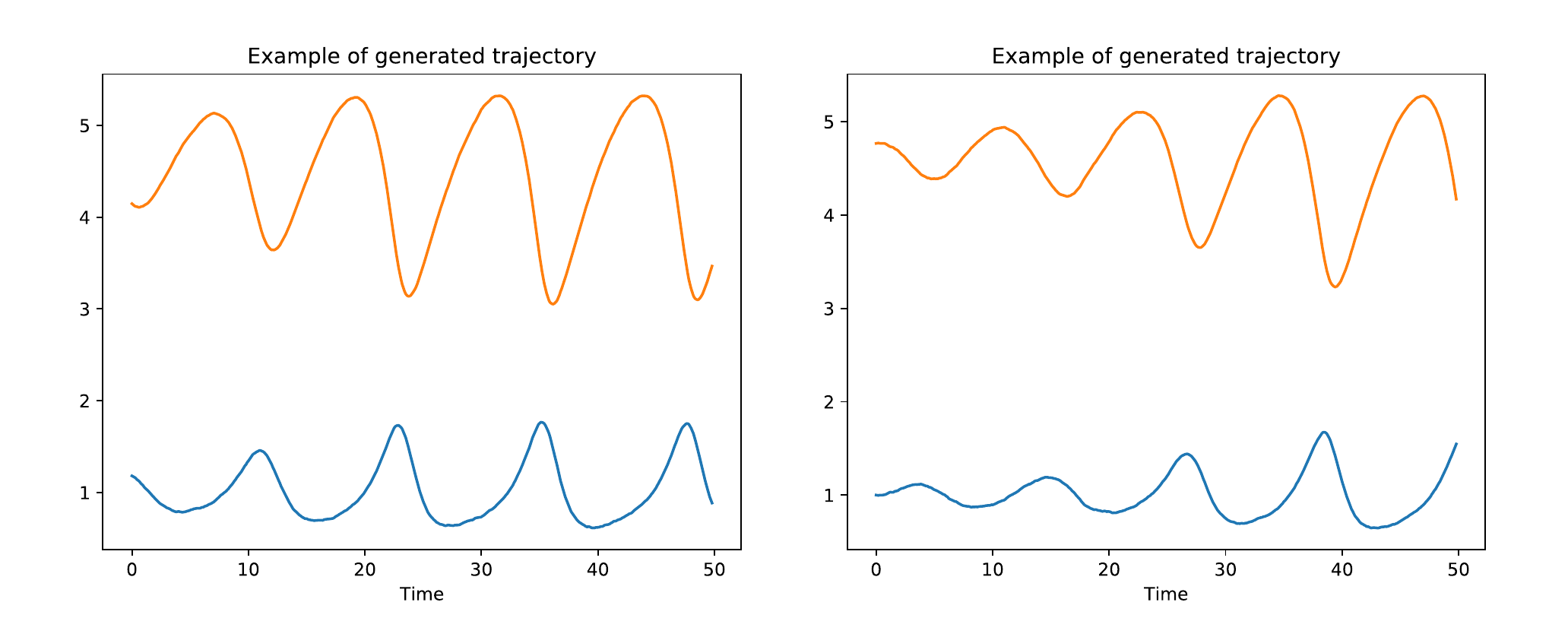}}
\caption{ Examples of generated trajectories for the stochastic Brusselator Eq.~\ref{eq:BXLator} over 50 seconds. Trajectories vary due to stochastic component and sensitivity to initial conditions. Orange and blue lines represent both components of the process.
}
\label{fig:BXLator_examples}
\end{figure}

Where $dW_1(t)$ and $dW_2(t)$ are correlated Brownian motions with correlation coefficient $\rho$. We simulate 1,000 trajectories driven by the dynamics given in Eq.~\ref{eq:BXLator} with parameters $a = 0.3$ and $b = 1.4$ such that the ODE is unstable. Figure \ref{fig:BXLator_examples} show some realization of this process. The data set we use for training consists in random samples from those trajectories of length 50. We sample sporadically with an average rate of 4 samples every 10 seconds.

\begin{figure}[hbt]
\centering
\centerline{\includegraphics[width=15cm]{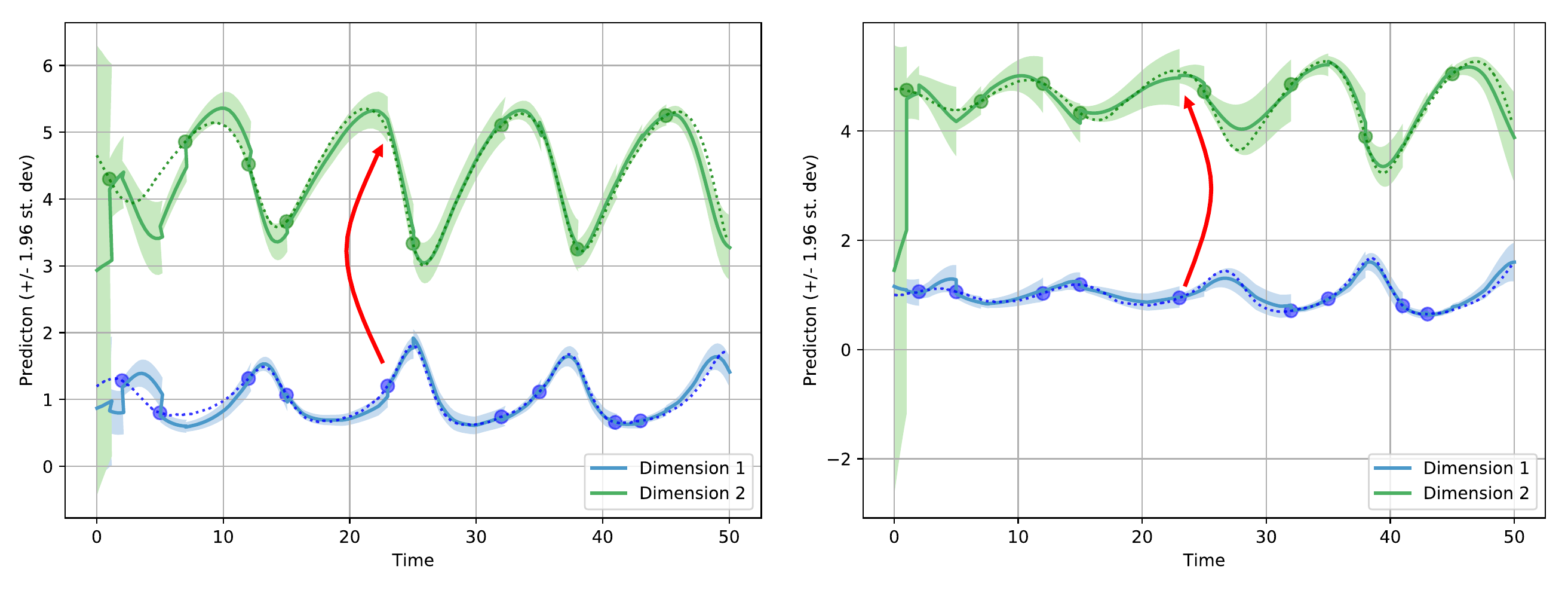}}
\caption{ Examples of predicted trajectories for the Brusselator. The model has been trained with DOPRI solver. Solid line shows the predicted \emph{filtered} mean, the shaded areas show the 95\% confidence interval while dotted lines represent the true generative process. The dots show the available observations for the filtering. Red arrows show the collapse of the belief function from one dimension to another.
}
\label{fig:BXLator_trained}
\end{figure}

Figures \ref{fig:BXLator_trained} show the predictions of the trained model on different samples of the proposed stochastic Brusselator process (newly generated samples). At each point in time are displayed the means and the standard deviation of the \emph{filtered} process. We stress that it means that those predictions only use the observations prior to them. Red arrows show that information is shared between both dimensions of the process. The model is able to pick up the correlation between dimensions to update its belief about one dimension when only the other is observed. The model presented in these figures used 50 dimensional latents with DOPRI solver.

\section{MIMIC-III: preprocessing details}
\label{app:mimic3}

MIMIC-III is a publicly available database containing deidentified health-related data associated for about 60,000 admissions of patients who stayed in critical care units of the Beth Israel Deaconess Medical Center between 2001 and 2012. To use the database, researchers must formally request access to the data via \url{http://mimic.physionet.org}. 

\subsection{Admission/Patient clean-up}

We only take a subset of admissions for our analysis. We select them on the following criteria:
\begin{itemize}
    \item Keep only patient who are in the metavision system.
    \item Keep only patients with single admission.
    \item Keep only patients whose admission is longer than 48 hours, but less than 30 days.
    \item Remove patients younger than 15 years old at admission time.
    \item Remove patients without chart events data.
    \item Remove patients with fewer than 50 measurements over the 48 hours. (This corresponds to measuring only half of retained variable a single time in 48 hours.)
\end{itemize}
 This process restricts the data set to 21,250 patients.
 
\subsection{Variables preprocessing}

The subset of 96 variables that we use in our study are shown in Table \ref{tab:MIMIC-feats}.
For each of those, we harmonize the units and drop the uncertain occurrences. We also remove outliers by discarding the measurements outside the 5 standard deviations interval. For models requiring binning of the time series, we map the measurements in 30-minute time bins, which gives 97 bins for 48 hours. When two observations fall in the same bin, they are either averaged or summed depending on the nature of the observation. Using the same taxonomy as in Table \ref{tab:MIMIC-feats}, lab measurements are averaged, while inputs, outputs, and prescriptions are summed. 

This gives a total of 3,082,224 unique measurements across all patients, or an average of 145 measurements per patient over 48 hours.

\begin{table}[htbp]
    \begin{adjustbox}{width=\textwidth}
   \begin{tabular}{| l | c | c | r | } 
     \hline
     \multicolumn{4}{|c|}{\textbf{Retained Features}} \\
     \hline
      Lab measurements   & Inputs & Outputs & Prescriptions\\
      \hline
      Anion Gap & Potassium Chloride & Stool Out Stool & D5W \\
      Bicarbonate & Calcium Gluconate & Urine Out Incontinent & Docusate Sodium \\
      Calcium, Total & Insulin - Regular & Ultrafiltrate Ultrafiltrate & Magnesium Sulfate \\
      Chloride & Heparin Sodium & Gastric Gastric Tube & Potassium Chloride \\
       Glucose & K Phos & Foley & Bisacodyl \\
       Magnesium & Sterile Water & Void & Humulin-R Insulin \\
       Phosphate & Gastric Meds & TF Residual & Aspirin \\
       Potassium & GT Flush &Pre-Admission & Sodium Chloride 0.9\%  Flush \\
       Sodium & LR & Chest Tube 1 & Metoprolol Tartrate \\
       Alkaline Phosphatase & Furosemide (Lasix) & OR EBL & \\
       Asparate Aminotransferase & Solution & Chest Tube 2 & \\
        Bilirubin, Total & Hydralazine & Fecal Bag & \\
       Urea Nitrogen & Midazolam (Versed) & Jackson Pratt 1 & \\
       Basophils & Lorazepam (Ativan) & Condom Cath & \\
       Eosinophils & PO Intake & & \\
       Hematocrit & Insulin - Humalog & & \\
       Hemoglobin & OR Crystalloid Intake & & \\
       Lymphocytes & Morphine Sulfate & & \\
       MCH & D5 1/2NS & & \\
       MCHC & Insulin - Glargine & & \\
       MCV & Metoprolol & & \\
       Monocytes & OR Cell Saver Intake & & \\
       Neutrophils & Dextrose 5\% & & \\
       Platelet Count & Norepinephrine & & \\
       RDW & Piggyback & & \\
       Red Blood Cells & Packed Red Blood Cells & & \\
       White Blood Cells & Phenylephrine & & \\
       PTT & Albumin 5\% & & \\
       Base Excess & Nitroglycerin & & \\
       Calculated Total CO2 & KCL (Bolus) & & \\
       Lactate & Magnesium Sulfate (Bolus) & & \\
       pCO2 & & & \\
       pH & & & \\
       pO2 & & & \\
       PT & & & \\
       Alanine Aminotransferase & & & \\
       Albumin & & & \\
       Specific Gravity & & & \\
      \bottomrule
   \end{tabular}
   \end{adjustbox}
    \caption{Retained longitudinal features in the intensive care case study.}
   \label{tab:MIMIC-feats}
\end{table}

\section{USHCN-Daily: preprocessing details}
\label{app:ushcn}

The United States Historical Climatology Network (USHCN) data set contains data from 1,218 centers scattered across the US. The data is publicly available and can be downloaded at the following address: \url{https://cdiac.ess-dive.lbl.gov/ftp/ushcn_daily/}. All states files contain daily measurements for 5 variables: precipitation, snowfall, snow depth, maximum temperature and minimum temperature. 

\subsection{Cleaning and subsampling}

We first remove all observations with a bad quality flag, then remove all centers that do not have observation before 1970 and after 2001. We then only keep the observations between 1950 and 2000. We subsample the remaining observations to keep on average 5\% of the observations of each center. Lastly, we select the last 4 years of the kept series to be used in the analysis.

This process leads to a data set with 1,114 centers, and a total of 386,068 unique observations (or an average of 346 observations per center, sporadically spread over 4 years).

\section{Small-sample regime: additional results}
\label{app:small-sample}
In the main text of the paper, we presented the results for the Mean Square Error (MSE) for the different data subsets of MIMIC. In Table \ref{table:mini-negll}, we present the negative log-likelihood results. They further illustrate that the continuity prior embedded in our \method{} strongly helps in the small-sample regime.

\begin{table}[h]
\caption{Vitals forecasting results on MIMIC-III (NegLL) - Low number of samples setting}
\label{table:mini-negll}
\vskip 0.15in
\begin{center}
\begin{small}
\begin{sc}
\begin{tabular}{lc|c|c}
\toprule
& \multicolumn{1}{c}{1,000 Patients} & \multicolumn{1}{c}{2,000 Patients} & \multicolumn{1}{c}{Full}  \\
\midrule
Model &  NegLL &  NegLL &  NegLL \\

\midrule

Neural-ODE       & $1.40 \pm 0.01$  & $1.39 \pm 0.005$  & $1.35 \pm 0.01$ \\ 
\hline
GRU-Disc-Bayes  & $1.35 \pm 0.01$  & $1.20 \pm 0.015$  & $\mathbf{0.74} \pm 0.04$ \\
\hline
GRU-ODE-Bayes     & $\mathbf{1.23} \pm 0.006$  & $\mathbf{1.13} \pm 0.01$  & $0.83 \pm 0.04$ \\

\bottomrule
\end{tabular}
\end{sc}
\end{small}
\end{center}
\vskip -0.1in
\end{table}

\section{Computing Infrastructure}
\label{app:infrastructure}

All models were run using a NVIDIA P100 GPU with 16GB RAM and 9 CPU cores (Intel(R) Xeon(R) Gold 6140). Implementation was done in Python, using Pytorch as autodifferentitation package. Required packages are available in the code

\section{Hyper-parameters used}
\label{app:hyper-params}

All methods were trained using the same dimension for the hidden $\mathbf{h}$, for sake of fairness. For each fold, we tuned the following hyper-parameters using a 20\% left out validation set:

\textbf{Dropout} rate of $0$, $0.1$, $0.2$ and $0.3$.

\textbf{Weight decay}: $0.1$, $0.03$, $0.01$, $0.003$, $0.001$, $0.0001$ and $0$.

\textbf{Learning rate} : $0.001$ and $0.0001$

Best model was selected using early stopping and performance were assessed by applying the best model on a held out test set (10\% of the total data). The different folds were reused for each compared model for sake of reproducibility and fair comparison. We performed 5-fold cross validation and present the test performance average and standard deviation in all tables.

\end{document}